\documentclass[letterpaper, 10 pt, conference]{ieeeconf}  

\IEEEoverridecommandlockouts                              
\usepackage{graphicx} 

\usepackage[margin=0pt,font=small,labelfont=bf]{caption}

\usepackage{amsmath,amssymb,amsfonts}
\usepackage{bm}

\usepackage{pifont}
\usepackage{stfloats}
\usepackage{subcaption}
\usepackage{lipsum}  
\usepackage{bm}
\usepackage{cite}
\usepackage{optidef}
\usepackage{booktabs} 

\usepackage{tabularx}
\usepackage{algorithmic}

\usepackage{booktabs}
\usepackage{multirow}
\usepackage{booktabs} 
\usepackage[colorlinks,bookmarksopen,bookmarksnumbered,linkcolor=blue,citecolor=blue,urlcolor=blue]{hyperref}
\usepackage{pbalance}

\usepackage[a-1b]{pdfx}

\title{Hierarchical Tri-manual Planning for Vision-assisted Fruit Harvesting \\with Quadrupedal Robots}
\author{Zhichao~Liu\textsuperscript{\textdagger,1},
        Jingzong~Zhou\textsuperscript{\textdagger,2},
        and~Konstantinos~Karydis\textsuperscript{2}
\thanks{\textsuperscript{\textdagger} Equal contribution. 
\textsuperscript{1} Institute for Integrative \& Innovative Research (I$^3$R), University of Arkansas, USA. {\tt\small zhichaol@uark.edu}. 
\textsuperscript{2} Department of Electrical and Computer Engineering, University of California, Riverside, USA.
{\tt\small \{jzhou227, karydis\}@ucr.edu}.}
\thanks{We gratefully acknowledge the support of NSF \#CMMI-2046270 and \#CMMI-2326309, USDA-NIFA \#2021-67022-33453, ONR \#W911NF-22-1-0156, and The University of California under grant UC-MRPI M21PR3417. 
Any opinions, findings, and conclusions or recommendations expressed in this material are those of the authors and do not necessarily reflect the views of the funding agencies.}
}

\begin{document}
%
\maketitle
\begin{abstract}
This paper addresses the challenge of developing a multi-arm quadrupedal robot capable of efficiently harvesting fruit in complex, natural environments. To overcome the inherent limitations of traditional bimanual manipulation, we introduce the first three-arm quadrupedal robot LocoHarv-3, that builds on top of the Spot quadruped, and propose a novel hierarchical tri-manual planning approach for automated fruit harvesting with collision-free trajectories between the built-in end-effector of Spot and our custom-made bimanual manipulator. Our comprehensive semi-autonomous framework integrates teleoperation, supported by LiDAR-based odometry and mapping, with learning-based visual perception for accurate fruit detection and pose estimation. Validation is conducted through a series of controlled indoor experiments using motion capture and extensive field tests in natural settings. Results demonstrate a 90\% success rate in in-lab settings with a single attempt, and field trials further verify the system's robustness and efficiency in more challenging real-world environments.
\end{abstract}

\IEEEpeerreviewmaketitle

\section{Introduction}
%
%
%
%
In precision agriculture, robotic arms offer notable technical benefits, including improved operational precision and efficiency, which hold potential to help reduce labor costs, and increase support for environmental sustainability~\cite{bac2014harvesting}. 
However, the complexity of agricultural environments and the demands of high-intensity production require robust and adaptable systems capable of handling crop variability~\cite{bechar2016agricultural}. 

Teleoperation systems for agricultural robots have witnessed widespread adoption~\cite{opiyo2021review}. 
Teleoperated robotic arms, in particular, can offer precise human-in-the-loop control for delicate tasks, versatility in handling various crops, enhanced safety by allowing remote operation in hazardous environments, and reduced physical labor for workers~\cite{murakami2008development}. 
Emerging technologies like augmented reality (AR) and virtual reality (VR) can further improve teleoperated robotic manipulation by enhancing user interaction and precision~\cite{chen2020real}. 
However, they often have high operational costs, rely on skilled operators, and require a steep learning curve for effective use. 
The dependency on human oversight also limits full automation and scalability in large-scale farming operations~\cite{fountas2020agricultural}. 

Agricultural robots utilizing multi-arm configurations have been gaining momentum. 
Allowing bimanual robots to perform coordinated actions with both arms, for instance, is crucial for achieving high proficiency~\cite{cohn2024constrained, amice2024certifying, chaki2024quadratic}. 
Dual-arm manipulators have also been successfully applied in agricultural settings. 
For instance, related works have addressed bimanual harvesting for aubergines~\cite{sepulveda2020robotic}, kiwifruit~\cite{he2022double}, and pears~\cite{yoshida2022automated}. 
In earlier work, we introduced a bimanual harvester with various end-effectors to address the elastic energy that may hinder the efficient harvesting of avocados~\cite{liu2024vision}. 
While dual-arm systems with a fixed base have a limited workspace, this can be enhanced by incorporating a third arm as the movable base with additional degrees of freedom (DoFs).

\begin{figure}[!t]
\vspace{6pt}
    \centering
    \includegraphics[width=0.96\columnwidth]{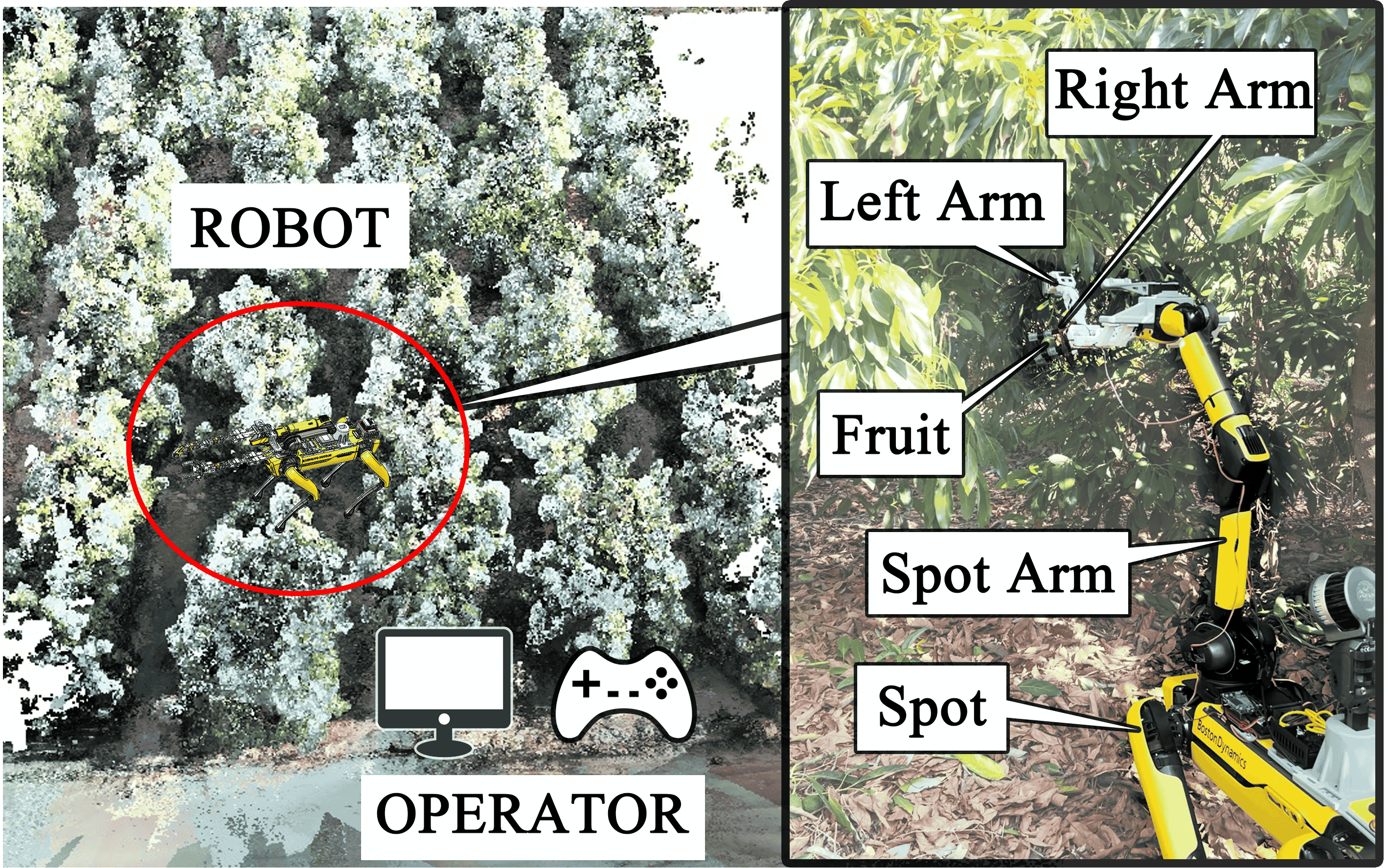}
    \vspace{-3pt}
    \caption{Illustration of semi-autonomous avocado harvesting using the proposed LocoHarv-3 platform and tri-manual path planning.}
    \label{fig:illustration_in_field}
    \vspace{-18pt}
\end{figure}


Mounting manipulators on mobile robotic platforms can extend the capabilities and improve harvesting efficiency~\cite{bechar2016agricultural}. 
Wheeled robots have been used to navigate flat or semi-structured terrains~\cite{williams2019robotic, dechemi2023robotic, chatziparaschis2024go,teng2025adaptive}. 
However, they often struggle with uneven or rough terrain, thereby limiting their use in rugged outdoor environments and tight or cluttered spaces because of limited maneuverability. 
Aerial robots equipped with robotic arms can enable interaction with objects from the air, adding a new dimension to robotic manipulation~\cite{kotarski2022toward, liu2024vision, gonzalez2024controlled}. 
However, aerial robots face limitations in payload capacity, operational time, energy efficiency, and stability. 

Legged robots (quadrupeds mostly) offer enhanced mobility on rough terrain (e.g., ANYmal~\cite{hutter2016anymal}, Boston Dynamics Spot, and Unitree B2). 
Despite their improved mobility and payload capacity, their use in agriculture is largely confined to applications like transportation, inspection, and mapping, with limited physical interaction~\cite{ferreira2022survey, tranzatto2022cerberus, zhang2024research}. 
Controlling robotic arms mounted on legged robots remains a challenge~\cite{chai2022survey, bertoncelli2020linear, polverini2020multi}. 
Existing solutions rarely address agricultural applications and are mostly limited to single-arm manipulation. 
To the best of the authors' knowledge, no work has yet addressed the challenge of deploying multi-arm quadrupedal robots for agricultural tasks.

In this work, we introduce an extended locomanipulation-capable robotic fruit harvester
that features three arms, called LocoHarv-3 (Fig.~\ref{fig:illustration_in_field}). 
Built on top of the Spot (with arm) quadruped~\cite{boston_dynamics_spot_arm}, the LocoHarv-3 features unique three-arm manipulation with various end-effectors specialized for agricultural tasks. 
The contributions extend to a first-of-its-kind hierarchical tri-manual planner to generate collision-free paths for dexterous and efficient fruit harvesting. 
Utilizing our prior work on vision-assisted bimanual autonomous harvester~\cite{liu2024vision}, this work develops a holistic framework for quadrupedal harvesting tasks in outdoor agricultural environments. 
Assisted with LiDAR SLAM, the framework comprises teleoperated navigation of the quadruped, hierarchical tri-manual planning and deep-learning-based fruit detection and pose estimation, and bimanual autonomous harvesting. 
Our approach is validated via a series of indoor and field experiments, demonstrating its potential to improve efficiency and scalability of autonomous agricultural systems.           


\section{LocoHarv-3 Platform}
\begin{figure}[!t]
    \vspace{2pt}
    \centering
        \includegraphics[width=0.8\columnwidth]{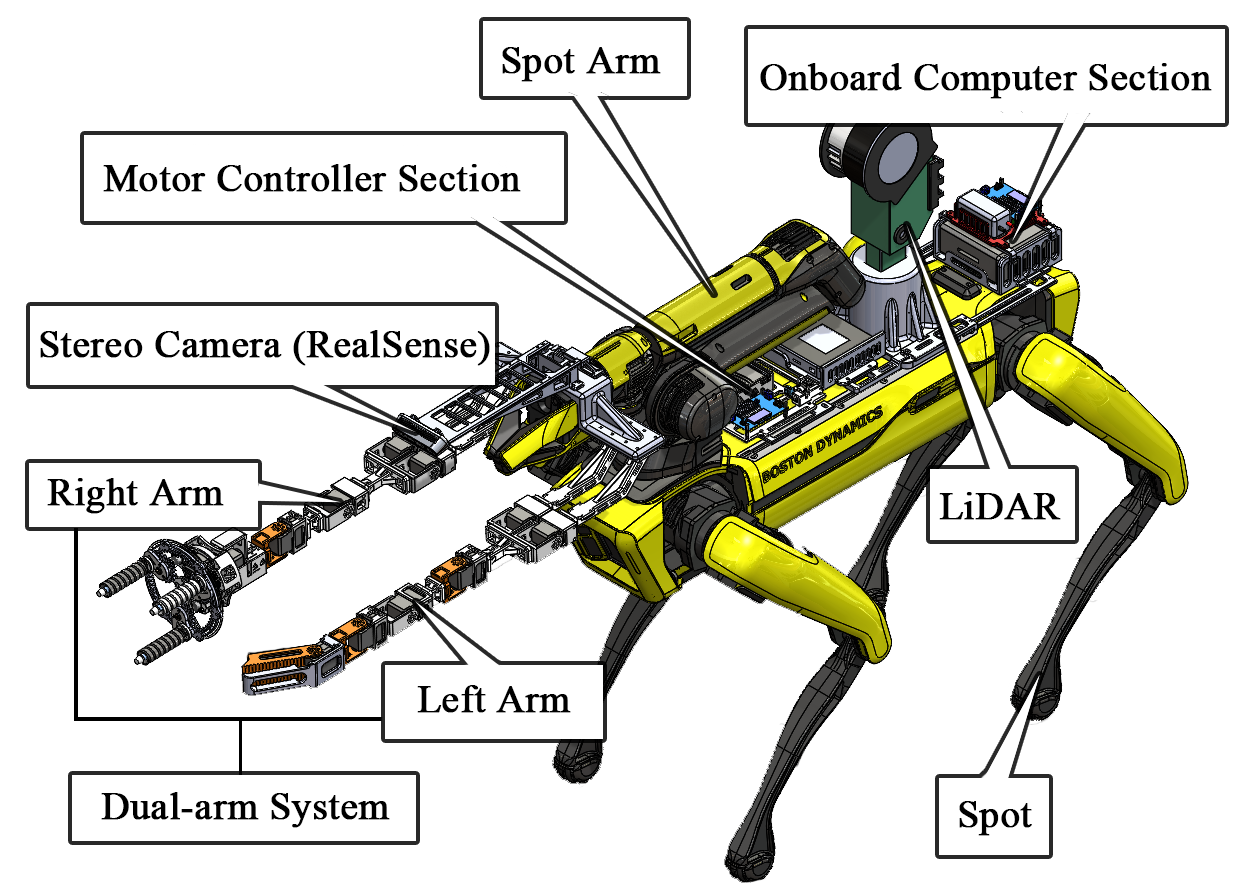} 
        \caption{The quadrupedal tri-manual platform LocoHarv-3.}
        \label{fig:first}
    \vspace{-10pt}
\end{figure}

We have retrofitted Spot's arm with a custom-built dual-arm system, attached on the former's wrist via a custom-built 3D-printed mounting base. 
Key hardware components are shown in Fig.~\ref{fig:first}. 
An onboard computer (Intel NUC with an i7-10710U processor, 16 GB RAM, and 512 GB SSD storage) is used for dual-arm motion planning and control as well as vision-based fruit detection and localization. 
A stereo camera (Intel RealSense D435i) is used for real-time scene understanding, while the motor controller section provides low-level control of the arms' motors (LX-16A bus servomotors). 
The motor controller section comprises a bus-linker, a 12V lithium-ion power bank, and a DC converter. 
The onboard computer and the stereo camera are powered by a 22.2V LiPo battery. 
DC converters provide the appropriate voltage (19V for mini PC and 7V for servomotors). 
A bus-linker processes control messages from the onboard computer to the servomotors. 
The LiDAR generates a map to localize LocoHarv-3 in the field in real-time.

The dual-arm assembly is primarily 3D-printed (Bambu Lab X1-Carbon Combo 3D printer) using carbon-fiber-reinforced material (PAHT-CF) for critical components and basic PLA material for non-critical components. 
The assembly comprises a left arm and a right arm. 
The role of the former is to hold the peduncle of the fruit, while the latter aims to engage with the fruit and harvest it; the need for both arms, especially in avocado picking, has been demonstrated in earlier work~\cite{zhou2024design,liu2024vision}. 
The left arm has four DoFs (excluding its end-effector), while the right arm has three DoFs (also excluding its end-effector). 
The weight of the dual-arm system is $1.63$\;kg.\footnote{~While the Spot arm can lift up to $11$\;kg and drag up to $25$\;kg, its end-effector (where our bimanual system is mounted) can hold $2.5$\;kg. 
This limits the weight of the dual-arm system and hence both left and right arms were designed to have fewer than 6 DoFs.}
All joints in both arms (excluding the two end-effectors) are revolute and are powered by either one or two servomotors. 
At the operating voltage of 7V, the LX-16A bus servo delivers a torque of about $1.7$ (\text{N·m}). 
To evenly distribute the load, 
the first joint of both left arm and right arm incorporates two bus servomotors operating in tandem. 

\section{Hierarchical Tri-manual Planning}

\begin{figure}[!t]
    \vspace{6pt}
    \centering
    \includegraphics[width=0.88\columnwidth]{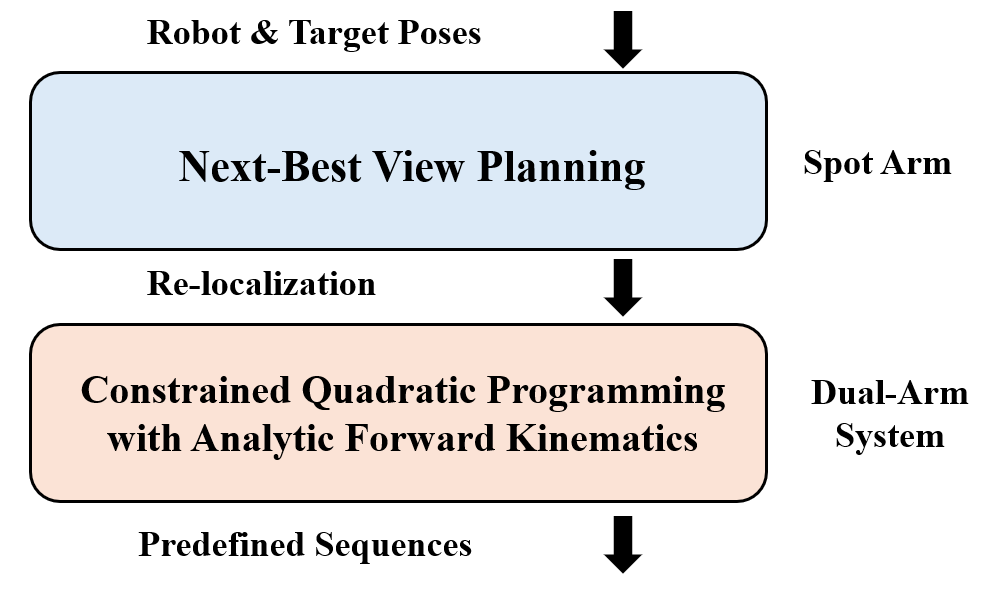}
    \vspace{-4pt}
    \caption{Overview of our hierarchical tri-manual planning approach.}
    \label{fig:overview}
    \vspace{-12pt}
\end{figure}

As demonstrated in our prior work~\cite{liu2024vision}, the dual-arm system has a constrained workspace, limited to the area between the two arms. 
Although quadrupedal platforms offer improved traversability and payload capacity, they lack the flexibility in 3D space compared to aerial systems. 
To address this limitation, we introduce a first-of-its-kind three-arm manipulation system, as well as a novel hierarchical tri-manual planning approach. 
Our approach (Fig.~\ref{fig:overview}) includes two parts: next-best view planning for the Spot arm and constrained quadratic programming with analytic forward kinematics for the dual-arm system.   

\subsection{Next-Best View Planning}

Next-best view (NBV) planning algorithms are commonly employed to enhance system perception and autonomy~\cite{bircher2016receding, dunn2009developing}. 
In addition to maximizing the information gained about the environment~\cite{breyer2022closed}, our NBV planning method strives to bridge the gap between the target and the workspace of the dual-arm system, while accounting for collision avoidance and orientation constraints. 
We follow the notation of~\cite{manipulation}; $\bm{q}_S$ denotes the joint values of the Spot's arm, while $\bm{X}_S$ and $\bm{X}^G_S$ denote the poses of the end-effector of the Spot's arm and the target. 
The forward kinematics of the Spot's arm is written as $f_{S, kin}$. 
The NBV planning can be formulated as       

\begin{mini}
    {\bm{q}_S}{|\bm{X}_S - \bm{X}^G_S|^2}{}{}
    \addConstraint{\bm{X}_S}{= f_{S,kin}(\bm{q}_S)}
    \addConstraint{|\bm{X}_S - \bm{X}^G_S|^2}{\geq d^\text{min}}
    \addConstraint{\bm{q}_{S [k]}}{\in [q^\text{min}_{S, [k]},\quad q^\text{max}_{S,[k]}]}
    \addConstraint{\bm{q}_S}{\ \text{collision free}}
    \label{eq:nbv}
\end{mini}
with $d^\text{min}$ a minimum distance from the target set empirically. 

Note that ``collision-free" operation includes both self-collision avoidance and obstacle avoidance. Self-collision avoidance is managed by the Spot's arm firmware~\cite{bdaiinstitute_spot_ros2}, while obstacle avoidance is ensured through the use of a 3D occupancy map. 
The pose $\bm{X}_S$ consists of both position $\bm{P}_S$ and orientation $\bm{\Phi}_S$, allowing the cost function to be expressed as $\bm{A}|\bm{P}_S - \bm{P}_S^G|^2 + \bm{B}|\bm{\Phi}_S - \bm{\Phi}_S^G|^2$, where $\bm{A}$ and $\bm{B}$ are tuning parameters. 
Our observations indicate that orientation plays a more significant role in the success of dual-arm manipulation, leading us to prioritize it in the cost function. 
To expedite computation, we fix the roll and pitch angles in $\bm{\Phi}^G_S$ during implementation. 
We also require that the Spot end-effector be horizontal to the ground when planning for the dual-arm system.

\subsection{Constrained QP with Analytic Forward Kinematics} \label{sec:opt}

We use $\bm{q}_i \ (i \in \{L,R\})$ to denote the joint values of the left and right arm, with $\bm{q}_i^0$ denoting the initial joint states. 
Similarly, $\bm{X}^0_i$ and $\bm{X}^G_i$ denote the initial and goal position of the end-effector for arm $i$. 
The forward and inverse kinematics of arm $i$ are denoted as $\bm{f}_{i, kin}$ and $\bm{f}^{-1}_{i, kin}$, respectively. 
Note that $\bm{f}_{L, kin} \in \mathbb{R}^{3 \times 6}$ and $\bm{f}_{R, kin} \in \mathbb{R}^{4 \times 6}$, and analytic solutions for inverse kinematics are unavailable. 
The path planning of the dual-arm system is formulated as a constrained quadratic programming (QP) problem, that is  
\begin{mini}
    {\bm{q}_i}{|\bm{q}_i - \bm{q}_i^0|^2}{}{}
    \addConstraint{\bm X_i^G}{= f_{i,kin}(\bm{q}_i)}
    \addConstraint{\bm{q}_{i [k]}}{\in [q^\text{min}_{i, [k]},\quad q^\text{max}_{i,[k]}]}
    \addConstraint{\bm{q}_i}{\ \text{collision free}}
    \label{eq:bimanual}
\end{mini}
where $q^\text{min}_{i, [k]}, q^\text{max}_{i, [k]}$ are limits of the joint $k$ of arm $i$. Several methods have been proposed to achieve self-collision avoidance of bimanual robots~\cite{cohn2024constrained, liu2018self, lei2020real} for redundant manipulators. 
However, due to the simplified structures of under-actuated arms, the project takes advantage of digital twins and utilizes the collision detection in MoveIt~\cite{moveit_ai} to reject colliding path candidates.      

We implement predefined sequences for efficient bimanual harvesting. 
They command the left arm's pincher end-effector to move and hold the peduncle while a rotary end-effector attached on the right arm encircles the target and rotates to detach it~\cite{zhou2024design, liu2024vision}. 
In detail, the left arm first moves toward its target pose $\bm{X}^G_L$ holding the peduncle to provide a relative fixed position for the right arm's target under the forward kinematic constraint $\bm{X_L^G}{= f_{L,kin}(\bm{q}_L)}$ and task-space constraint $\bm{X}^G_L=T(\bm{X}^G_R)$. 
Then the right arm approaches $\bm{X}^G_R$ under the forward kinematic constraints $\bm{X_R^G}{= f_{R,kin}(\bm{q}_R)}$ holding the target fruit and rotates by certain degrees to disengage the target from the peduncle.\footnote{~Note that the forward kinematic constraint is nonlinear. Study of the convexity of such problem is beyond the topic of this paper.} 
Problem~\eqref{eq:nbv} is responsible for the Spot arm. Problem \eqref{eq:bimanual} is responsible for left and right arms and the rates of sending the control commands while solving it is at $10$ Hz. 
Because of the limited DoFs of each arm, to guarantee success rate, the current planning target is under a predetermined pose for both left and right arm and it is set manually in the simulation environment. From the visual perception, if the detected target is close to this ideal target, the grasping sequence will then be conducted. 

\section{Learning-based Visual Perception} \label{sec:visual}

In line with our previous work~\cite{liu2024vision}, data from the stereo camera mounted on the Spot's arm gripper are integrated into a deep learning-based visual perception framework for detecting fruits and determining their location relative to the camera frame. 
While this section provides a brief overview, a detailed description can be found in~\cite{liu2024vision}.

\begin{figure}[ht]
    \vspace{-4pt}
    \centering
    \includegraphics[width=0.98\columnwidth]{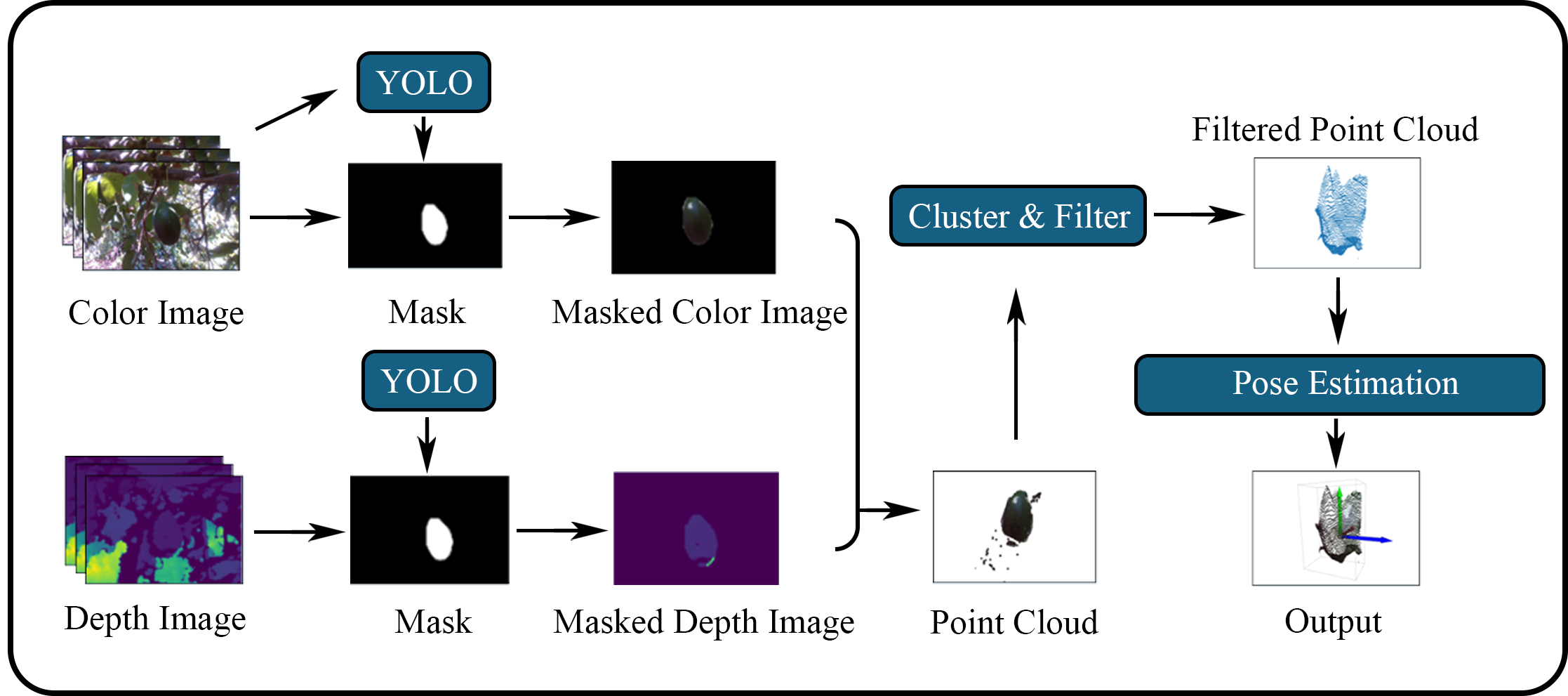}
    \vspace{0pt}
    \caption{Workflow of the learning-based fruit detection and pose estimation using an RGB-D camera.}
    \label{fig:perception}
    \vspace{-10pt}
\end{figure}

As depicted in Fig.~\ref{fig:perception}, the system utilizes a fine-tuned YOLOv8 framework~\cite{ultralytics2023yolov8} to detect fruits (in this case, avocados) in color images, segmenting them into 2D masks. 
Since the camera's color and depth sensors have different fields of view, depth images are preprocessed to align with the color images. 
The YOLO-generated mask is then applied to both the color and aligned depth images, enabling the creation of 3D point clouds for the segmented avocados. 
These point clouds are used to identify suitable avocado candidates by generating 3D bounding boxes. 
Given that the point clouds may contain multiple candidates as well as noise, histogram filtering is applied to organize the points by their distance from the camera. 
The number of detections from the YOLO model is used to set the number of clusters. 
Then, the geometric center of each 3D point cluster is determined as the location of the detected fruit, with the orientation computed via the bounding box method in Open3D~\cite{open3d}.

\section{Quadrupedal Fruit Harvesting Framework Assisted with LiDAR SLAM}\label{sec:framework}

\begin{figure}[!t]
    \vspace{6pt}
    \centering
    \includegraphics[width=0.7\columnwidth]{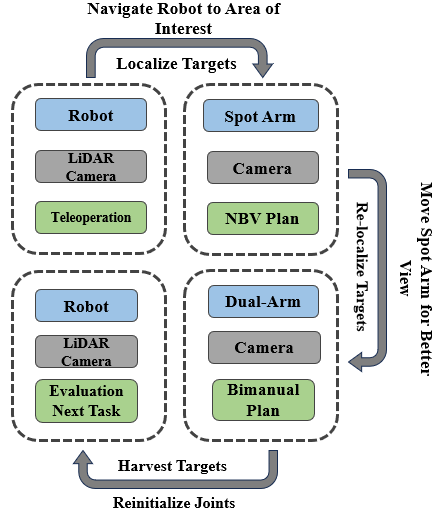}
    \vspace{0pt}
    \caption{Flowchart of the semi-autonomous fruit harvesting framework with the hierarchical tri-manual planning, learning-based visual perception, and LiDAR-SLAM-assisted teleoperation. }
    \label{fig:loop}
    \vspace{-18pt}
\end{figure}

Our overall framework is depicted in Fig.~\ref{fig:loop}. 
The robot first navigates to an area of interest, where it uses a 3D LiDAR for mapping and self-localization, as well as an RGB-D camera to localize targets, such as fruits, through teleoperation. 
Once the targets are initially identified by the camera mounted on the Spot arm, it is deployed to obtain a better view using the Next-Best View (NBV) planning algorithm~\eqref{eq:nbv}. 
After repositioning the arm for an improved view, the visual perception module re-localizes the targets to refine their exact positions. 
Following this, the custom dual-arm system is activated, and the bimanual planning strategy~\eqref{eq:bimanual} is executed, allowing the robot to manipulate both arms simultaneously for precise and coordinated harvesting of the targets.  
The two optimization problems are solved separately using the MoveIt RRT module. 
After each execution, the robot evaluates the next task and prepares for the next target by reinitializing its joints. 
The framework alternates between localization, arm adjustment, and dexterous manipulation, combining autonomous planning with some teleoperated oversight to achieve effective target harvesting.

\begin{figure*}[!t]
    \vspace{6pt}
    \centering
    \begin{minipage}[b]{0.9\textwidth}
        \centering
        \begin{subfigure}[b]{0.95\textwidth}
            \includegraphics[width=\textwidth]{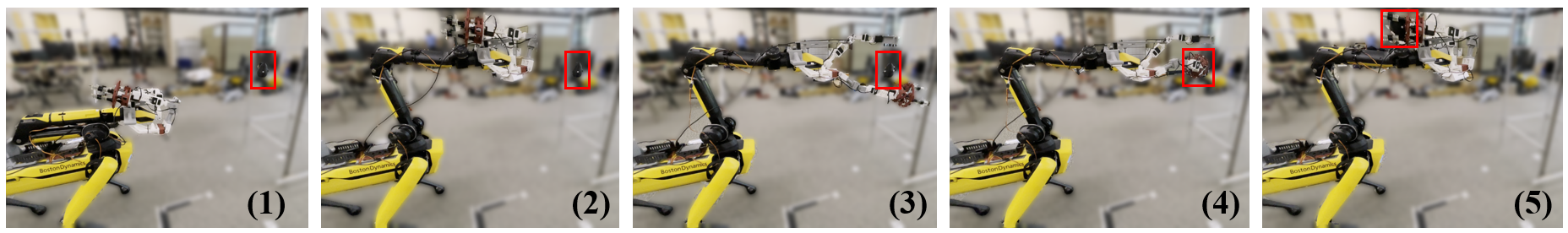}
            \caption{Fully automated harvesting manipulation in a controlled laboratory setting.}
            \label{fig:sequence}
        \end{subfigure}
        \vspace{2pt}
        \begin{subfigure}[b]{0.95\textwidth}
            \includegraphics[width=\textwidth]{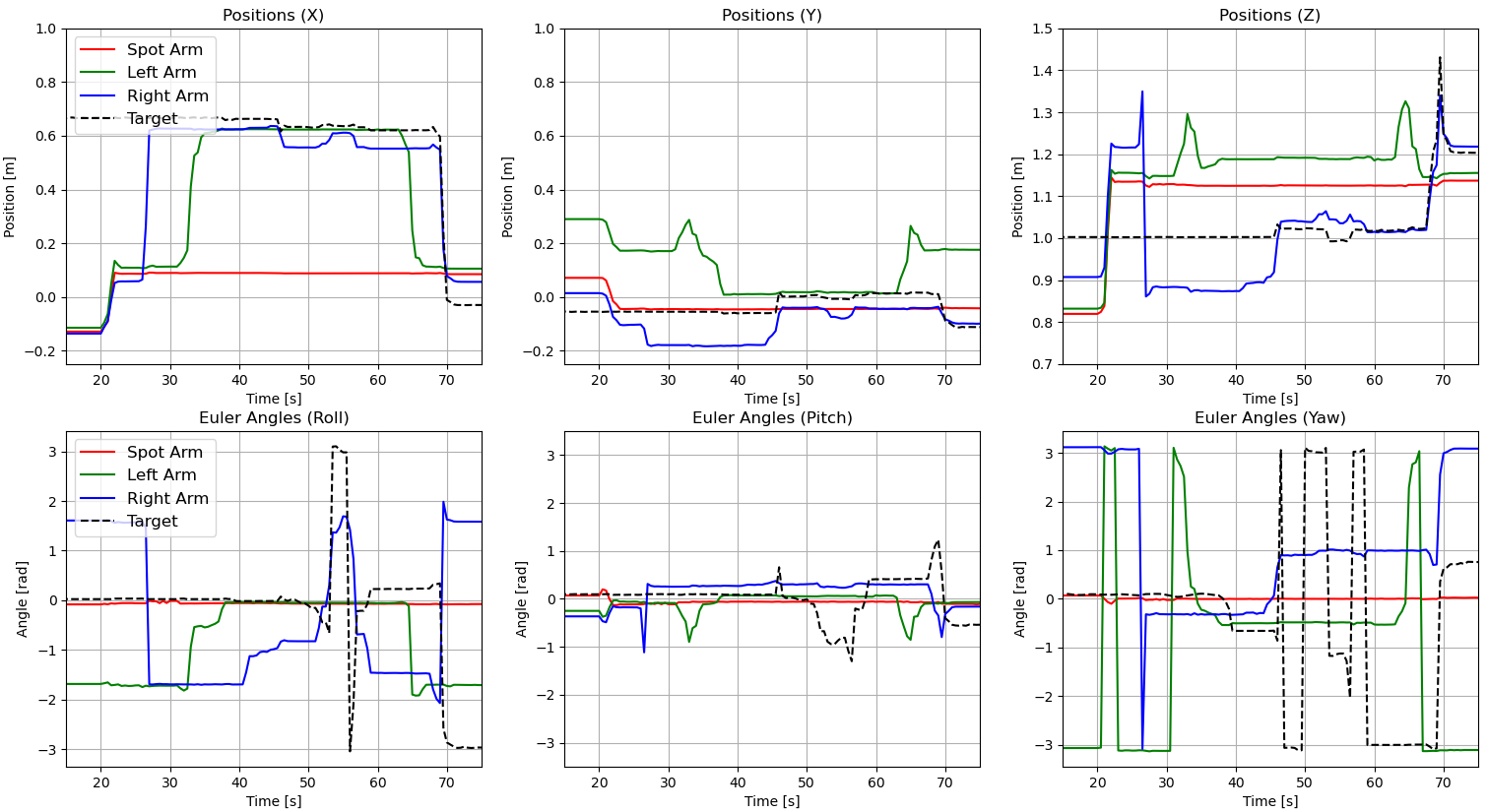}
            \caption{Positions and orientations.}
            \label{fig:position}
        \end{subfigure}
        \vspace{2pt}
        \begin{subfigure}[b]{0.92\textwidth}
            \includegraphics[width=\textwidth]{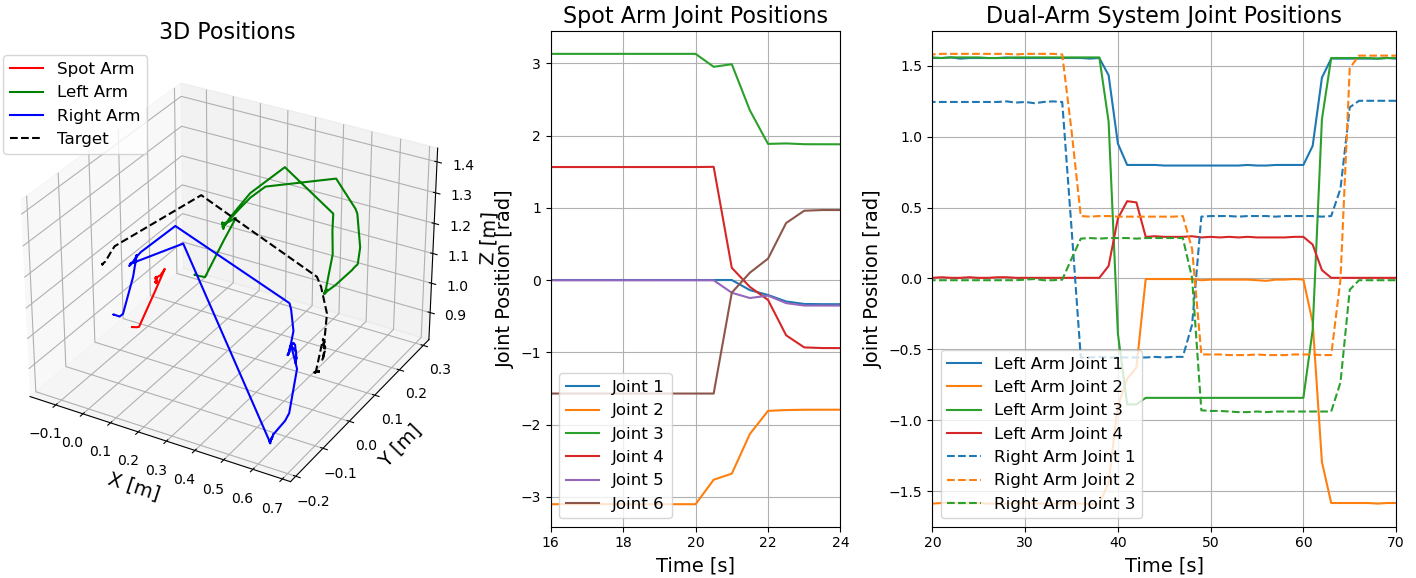}
            \caption{3D positions and joint values.}
            \label{fig:joint_value1}
        \end{subfigure}
    \end{minipage}
    \vspace{-5pt}
    \caption{Overview of indoor mockup-harvesting testing results. (a) Snapshots showed the robot reaching for and retrieving an artificial avocado suspended by a flexible cord, manipulation with related joint trajectories and 3D position and orientation trajectories. (b) and (c) Temporal evolution of state and configuration components during one of the indoor experimental trials. (Best viewed in color.)} 
    \vspace{-15pt}
    \label{fig:result1}
\end{figure*}

\section{Experiments}
The experimental analysis consists of two stages: indoor and outdoor testing. 
First, we conduct an in-depth examination of the tree-arm manipulation system in a controlled laboratory environment, where the robot is tasked with retrieving an artificial avocado. 
For precise tracking of position and orientation, we employ an OptiTrack motion capture system. 
In the second phase, we transition to field testing by deploying the robot at the Agricultural Experimental Station (AES) facilities at UC Riverside. 
Here, we evaluate the robot's performance in a real-world agricultural setting, focusing on the holistic semi-autonomous fruit harvesting framework. 
Outdoor experiments allow us to assess the robustness of our system under natural environmental conditions, ensuring the practical utility of the proposed solution.  

\subsection{Indoor Testing}
In the first experiment, the pose ground truth of the target is directly fed into the tri-manual planning algorithm; the visual perception module is not involved. 
The artificial fruit is suspended in mid-air with a flexible cord (area highlighted with a red box in Fig.~\ref{fig:sequence}), allowing it to swing freely to emulate the equivalent natural situation. 
The LocoHarv-3 robot is standing still with self-righting in front of the target. 
The dual-arm system is initially folded and pointing to the rear, as in Fig.~\ref{fig:sequence}(1), to minimize the impact on locomotion. 

After receiving the pose information of the target via motion capture, the robot determines a waypoint $\bm{X}_S$ for the Spot arm with an improved view by solving the optimization~\eqref{eq:nbv}. 
The desired waypoint is then inputted to the Spot arm, which is commanded to reach the position shown in Fig.~\ref{fig:sequence}(2). 
The algorithm extends to the second stage of planning by executing constrained quadratic programming with analytic forward kinematics, following a predefined sequence to harvest the target using both arms and end-effectors. 
The left arm reaches position ${X_L^G}$ and holds the peduncle (Fig.~\ref{fig:sequence}(3)). 
The right arm then follows its path to reach and grasp the target (Fig.~\ref{fig:sequence}(4)). 
The end-effector of the right arm rotates to detach the target from the peduncle, while the left arm reduces the impact of the elastic energy from the swinging peduncle. 
After harvesting the target, both arms return to their initial folded states (Fig.~\ref{fig:sequence}(5)). 

We present the positions and Euler angles with rotation sequence ZYX of the three end-effectors, along with the target, in Fig.~\ref{fig:position}. 
As depicted, the $X$, $Y$, and $Z$ positions around the 20-sec mark indicate that the Spot arm effectively tracks the output of the NBV planning, with noticeable displacements in all three end-effectors. 
Benefiting from its robust motors, the Spot arm requires approximately $3$\;sec to reach the desired waypoint. 
It can be verified by observing the Euler angles in Fig.~\ref{fig:position} that the NBV planning prioritizes the alignment of the roll and pitch angles, while the yaw angle of the target remains uncontrolled and can vary. 
Prior to initiating the second stage of the tri-manual planning process at around the 40-sec mark, both the left and right arms follow empirically predefined trajectories to extend horizontally. 
This prepares the system for the subsequent bimanual manipulation, which continues until approximately $25$\;sec. 
During the harvesting process, the roll angles of both the target and the right arm's end-effector exhibit similar movements, although the target displays a larger angle due to relative motion. 
Upon completing the task, both arms return to their initial folded positions, successfully retrieving the target. 
This sequence highlights the precision and coordination involved in the tri-manual planning framework. 
We visualize the 3D positions of the three end-effectors and the target in Fig.~\ref{fig:joint_value1}, which illustrates the closed path of the dual-arm system as it returns with the harvested target. 

The joint values of both the Spot arm and the dual-arm system are visualized in Fig.~\ref{fig:joint_value1}. 
The specifications for the Spot arm can be found in~\cite{boston_dynamics_spot_arm}, while the joint specifications for the dual-arm system are detailed in our previous work~\cite{liu2024vision}. 
As illustrated in the figure, the constrained quadratic programming method, combined with analytic forward kinematics, enables the two under-actuated robotic arms to perform dexterous maneuvers and efficiently complete the fruit harvesting task.

To evaluate the efficiency and robustness of the tri-manual manipulation system, we instructed the robot to execute the framework continuously for 10 iterations and recorded the success rate for retrieving the target with a single attempt. 
A success rate of 90\% was observed, with only one failure attributed to the target slipping inside the gripper (which can be directly addressed in future work by increasing the friction inside the gripper using some padding for instance). 
This result demonstrates the robustness of the tri-manual manipulation framework for fruit harvesting.

\subsection{Outdoor Testing}
Outdoor testing aims to showcase the efficiency of the holistic framework to semi-autonomously harvest fruits (avocado in this test) by utilizing the novel three-arm quadrupedal robot, proposed hierarchical tri-manual planning method and the learning-based visual perception.

\begin{figure}[!t]
    \vspace{7pt}
    \centering
    \includegraphics[trim={0cm, 0cm, 0cm, 4cm},clip,width=0.88\columnwidth]{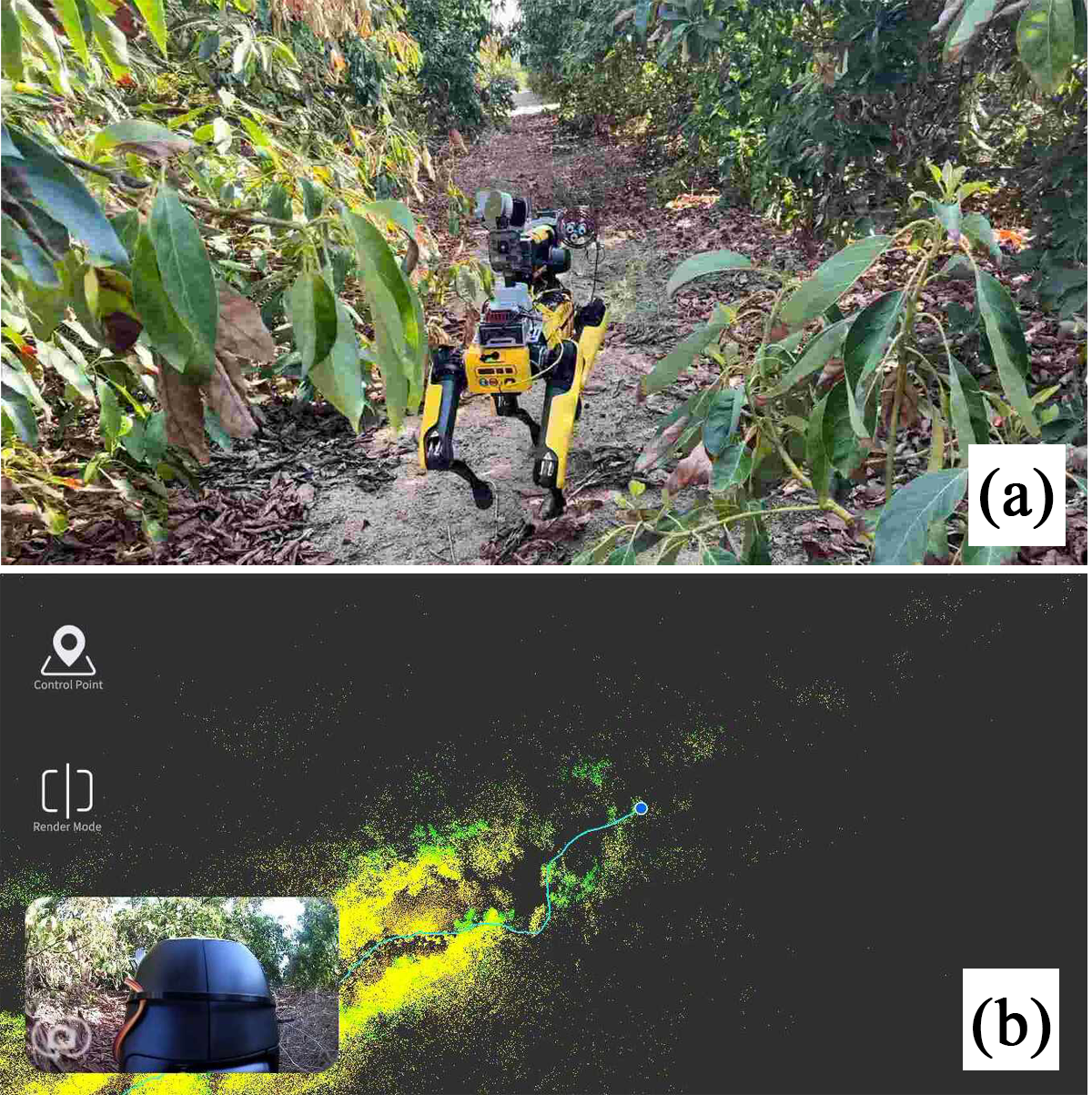}
    \vspace{0pt}
    \caption{(a) Snapshot shown the robot traversing rough terrains in AES fields at UC Riverside while carrying a heavy payload. (b) LiDAR-based mapping and localization assisting with teleoperation.}
    \label{fig:outdoor_lidar_map}
    \vspace{-18pt}
\end{figure}

As shown in Fig.~\ref{fig:outdoor_lidar_map}, the robot's localization is assisted by the Xgrids Lixel L2\cite{xgrids_lixell2}, which is equipped with a 32-beam LiDAR and runs SLAM algorithms. 
Meanwhile, mapping with colorized point clouds can create a digital twin of the agricultural field for high-fidelity simulation~\cite{teng2023multimodal}. 
The robot is tele-operated by an operator, who has access to the LiDAR-based localization, as well as images from Spot's body and arm cameras. 
The robot is capable of traversing highly challenging terrain and navigating through cluttered pathways in the AES field (Fig.~\ref{fig:outdoor_lidar_map}(a)). 

We evaluate the performance of visual detection and harvesting of real avocados in natural environments. 
After teleoperating the robot to approach the avocado tree, the robot initiates the NBV planning and follows the generated trajectory to ensure that the target avocados fall within the feasible subset of the robot's reachable workspace. 
Upon reaching the desired pose from the NBV planner, the learning-based visual perception system is activated using the RGB-D camera to perform target detection and pose estimation. 
If the target is detected outside the dual-arm system's workspace, a message is sent to the operator, prompting to teleoperate the robot and re-initiate the framework. 

\begin{figure}[!t]
\vspace{6pt}
    \centering
    \includegraphics[width=0.99\columnwidth]{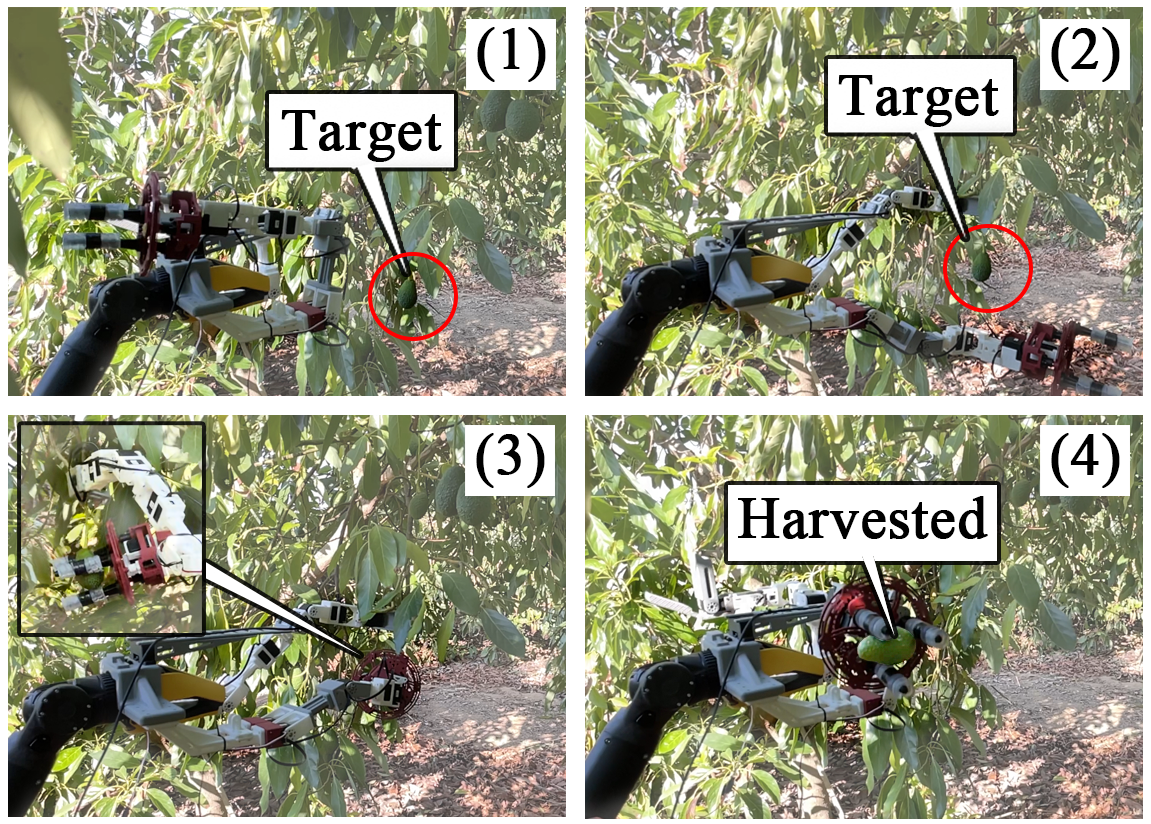}
    \vspace{-12pt}
    \caption{Complete manipulation for harvesting in the field: (1) Target being detected by the camera. (2) Left arm activated to grasp the peduncle to provide a stable posture of the target. (3) Right arm activated to harvest the target. (4) Target being harvested.}
    \label{fig:manipulation_in_field}
    \vspace{-6pt}
\end{figure}

Figure~\ref{fig:manipulation_in_field} illustrates a sequence of successfully retrieving an avocado. 
We also run 10 trials and recorded the success rate in a single attempt. 
Table~\ref{table:harvesting_success_rates_in_two_environments} contains the results. 
We observe a drop in the success rate in the natural environment compared to the controlled laboratory settings. 
Several factors may account for this difference. 
The pose estimation provided by the camera is subject to noise, which is further exacerbated by the movements of the Spot robot and its arm.
Ambient wind in the orchard causes the target avocados to shake, presenting additional challenges for both visual detection and precise grasping. 
However, the success rate test only accounts for a single harvesting attempt, and the overall efficiency can be significantly improved if multiple attempts are considered and re-run the framework.

\begin{table}[h!]
\centering
\caption{Harvesting Success Rates for Two Environments}
\label{table:harvesting_success_rates_in_two_environments}
\begin{tabular}{c | c}
\toprule
Environment & Success Rate $\uparrow$ \\
\midrule
Indoor & 90\% \\
Outdoor & 40\% \\
\bottomrule
\end{tabular}
\vspace{-18pt}
\end{table}

\section{Conclusion}
In this paper, we address the challenge of developing a multi-arm quadrupedal agricultural robot capable of harvesting fruits while navigating through challenging natural environments. To overcome the constrained workspace of bimanual manipulation, we introduce the first-of-its-kind three-arm quadrupedal robot and propose a novel hierarchical tri-manual planning approach that automates fruit harvesting with collision-free trajectories. We present a holistic semi-autonomous framework that integrates teleoperation, assisted by LiDAR-based odometry and mapping, with learning-based visual perception for fruit detection and pose estimation. Our approach is validated through a series of indoor experiments using motion capture and outdoor tests in natural environments. The results show that our method achieves a 90\% success rate with a single attempt in laboratory conditions. Field tests further validate the efficiency of the approach in challenging natural environments.

This works enables several promising future directions for research. 
Our approach can be tested with a wider range of fruits and vegetables. 
We also aim to integrate LiDAR sensors to enhance detection and pose estimation accuracy. 
Further, we intend to investigate multi-modal harvesting by incorporating wheeled, quadrupedal, and aerial harvesters. 
Finally, we seek to investigate soft-robotics-based solution, including harvesters with compliant end-effectors~\cite{liu2022safely} as well as soft legged platforms~\cite{liu2020sorx, liu2021position}.

\section*{Acknowledgment}
The authors would like to thank Cody Simons and Aritra Samanta for their assistance with the experiments.

\bibliographystyle{IEEEtran}
\bibliography{ref}

\begin{thebibliography}{10}
\providecommand{\url}[1]{#1}
\csname url@samestyle\endcsname
\providecommand{\newblock}{\relax}
\providecommand{\bibinfo}[2]{#2}
\providecommand{\BIBentrySTDinterwordspacing}{\spaceskip=0pt\relax}
\providecommand{\BIBentryALTinterwordstretchfactor}{4}
\providecommand{\BIBentryALTinterwordspacing}{\spaceskip=\fontdimen2\font plus
\BIBentryALTinterwordstretchfactor\fontdimen3\font minus \fontdimen4\font\relax}
\providecommand{\BIBforeignlanguage}[2]{{%
\expandafter\ifx\csname l@#1\endcsname\relax
\typeout{** WARNING: IEEEtran.bst: No hyphenation pattern has been}%
\typeout{** loaded for the language `#1'. Using the pattern for}%
\typeout{** the default language instead.}%
\else
\language=\csname l@#1\endcsname
\fi
#2}}
\providecommand{\BIBdecl}{\relax}
\BIBdecl

\bibitem{bac2014harvesting}
C.~W. Bac, E.~J. Van~Henten, J.~Hemming, and Y.~Edan, ``Harvesting robots for high-value crops: State-of-the-art review and challenges ahead,'' \emph{Journal of field robotics}, vol.~31, no.~6, pp. 888--911, 2014.

\bibitem{bechar2016agricultural}
A.~Bechar and C.~Vigneault, ``Agricultural robots for field operations: Concepts and components,'' \emph{Biosystems Engineering}, vol. 149, pp. 94--111, 2016.

\bibitem{opiyo2021review}
S.~Opiyo, J.~Zhou, E.~Mwangi, W.~Kai, and I.~Sunusi, ``A review on teleoperation of mobile ground robots: Architecture and situation awareness,'' \emph{International Journal of Control, Automation and Systems}, vol.~19, pp. 1384--1407, 2021.

\bibitem{murakami2008development}
N.~Murakami, A.~Ito, J.~D. Will, M.~Steffen, K.~Inoue, K.~Kita, and S.~Miyaura, ``Development of a teleoperation system for agricultural vehicles,'' \emph{Computers and Electronics in Agriculture}, vol.~63, no.~1, pp. 81--88, 2008.

\bibitem{chen2020real}
Y.~Chen, B.~Zhang, J.~Zhou, and K.~Wang, ``Real-time 3d unstructured environment reconstruction utilizing vr and kinect-based immersive teleoperation for agricultural field robots,'' \emph{Computers and Electronics in Agriculture}, vol. 175, p. 105579, 2020.

\bibitem{fountas2020agricultural}
S.~Fountas, N.~Mylonas, I.~Malounas, E.~Rodias, C.~Hellmann~Santos, and E.~Pekkeriet, ``Agricultural robotics for field operations,'' \emph{Sensors}, vol.~20, no.~9, p. 2672, 2020.

\bibitem{cohn2024constrained}
T.~Cohn, S.~Shaw, M.~Simchowitz, and R.~Tedrake, ``Constrained bimanual planning with analytic inverse kinematics,'' in \emph{IEEE International Conference on Robotics and Automation (ICRA)}, 2024, pp. 6935--6942.

\bibitem{amice2024certifying}
A.~Amice, P.~Werner, and R.~Tedrake, ``Certifying bimanual rrt motion plans in a second,'' in \emph{IEEE International Conference on Robotics and Automation (ICRA)}, 2024, pp. 9293--9299.

\bibitem{chaki2024quadratic}
T.~Chaki and T.~Kawakami, ``Quadratic programming based inverse kinematics for precise bimanual manipulation,'' in \emph{IEEE International Conference on Robotics and Automation (ICRA)}, 2024, pp. 16\,024--16\,030.

\bibitem{sepulveda2020robotic}
D.~Sep{\'u}Lveda, R.~Fern{\'a}ndez, E.~Navas, M.~Armada, and P.~Gonz{\'a}lez-De-Santos, ``Robotic aubergine harvesting using dual-arm manipulation,'' \emph{IEEE Access}, vol.~8, pp. 121\,889--121\,904, 2020.

\bibitem{he2022double}
Z.~He, L.~Ma, Y.~Wang, Y.~Wei, X.~Ding, K.~Li, and Y.~Cui, ``Double-arm cooperation and implementing for harvesting kiwifruit,'' \emph{Agriculture}, vol.~12, no.~11, p. 1763, 2022.

\bibitem{yoshida2022automated}
T.~Yoshida, Y.~Onishi, T.~Kawahara, and T.~Fukao, ``Automated harvesting by a dual-arm fruit harvesting robot,'' \emph{ROBOMECH journal}, vol.~9, no.~1, p.~19, 2022.

\bibitem{liu2024vision}
Z.~Liu, J.~Zhou, C.~Mucchiani, and K.~Karydis, ``Vision-assisted avocado harvesting with aerial bimanual manipulation,'' \emph{arXiv preprint arXiv:2408.09058}, 2024.

\bibitem{williams2019robotic}
H.~A. Williams, M.~H. Jones, M.~Nejati, M.~J. Seabright, J.~Bell, N.~D. Penhall, J.~J. Barnett, M.~D. Duke, A.~J. Scarfe, H.~S. Ahn \emph{et~al.}, ``Robotic kiwifruit harvesting using machine vision, convolutional neural networks, and robotic arms,'' \emph{biosystems engineering}, vol. 181, pp. 140--156, 2019.

\bibitem{dechemi2023robotic}
A.~Dechemi, D.~Chatziparaschis, J.~Chen, M.~Campbell, A.~Shamshirgaran, C.~Mucchiani, A.~Roy-Chowdhury, S.~Carpin, and K.~Karydis, ``Robotic assessment of a crop’s need for watering: Automating a time-consuming task to support sustainable agriculture,'' \emph{IEEE Robotics \& Automation Magazine}, 2023.

\bibitem{chatziparaschis2024go}
D.~Chatziparaschis, H.~Teng, Y.~Wang, P.~Peiris, E.~Seudiero, and K.~Karydis, ``On-the-go tree detection and geometric traits estimation with ground mobile robots in fruit tree groves,'' in \emph{IEEE International Conference on Robotics and Automation (ICRA)}, 2024, pp. 15\,840--15\,846.

\bibitem{teng2025adaptive}
H.~Teng, Y.~Wang, D.~Chatziparaschis, and K.~Karydis, ``Adaptive lidar odometry and mapping for autonomous agricultural mobile robots in unmanned farms,'' \emph{Computers and Electronics in Agriculture}, vol. 232, p. 110023, 2025.

\bibitem{kotarski2022toward}
D.~Kotarski, P.~Piljek, M.~Pranjic, and J.~Kasac, ``Toward modular aerial robotic system for applications in precision agriculture,'' in \emph{IEEE International Conference on Unmanned Aircraft Systems (ICUAS)}, 2022, pp. 1530--1537.

\bibitem{gonzalez2024controlled}
A.~Gonz{\'a}lez-Morgado, E.~Cuniato, M.~Tognon, G.~Heredia, R.~Siegwart, and A.~Ollero, ``Controlled shaking of trees with an aerial manipulator,'' \emph{IEEE/ASME Transactions on Mechatronics}, 2024.

\bibitem{hutter2016anymal}
M.~Hutter, C.~Gehring, D.~Jud, A.~Lauber, C.~D. Bellicoso, V.~Tsounis, J.~Hwangbo, K.~Bodie, P.~Fankhauser, M.~Bloesch \emph{et~al.}, ``Anymal-a highly mobile and dynamic quadrupedal robot,'' in \emph{IEEE/RSJ international conference on intelligent robots and systems (IROS)}, 2016, pp. 38--44.

\bibitem{ferreira2022survey}
J.~Ferreira, A.~P. Moreira, M.~Silva, and F.~Santos, ``A survey on localization, mapping, and trajectory planning for quadruped robots in vineyards,'' in \emph{IEEE International Conference on Autonomous Robot Systems and Competitions (ICARSC)}, 2022, pp. 237--242.

\bibitem{tranzatto2022cerberus}
M.~Tranzatto, T.~Miki, M.~Dharmadhikari, L.~Bernreiter, M.~Kulkarni, F.~Mascarich, O.~Andersson, S.~Khattak, M.~Hutter, R.~Siegwart \emph{et~al.}, ``Cerberus in the darpa subterranean challenge,'' \emph{Science Robotics}, vol.~7, no.~66, p. eabp9742, 2022.

\bibitem{zhang2024research}
J.~Zhang, X.~Wang, and L.~Zheng, ``Research on autonomous navigation system of agricultural quadruped robot,'' in \emph{IEEE International Conference on Mechatronics and Automation (ICMA)}, 2024, pp. 1286--1290.

\bibitem{chai2022survey}
H.~Chai, Y.~Li, R.~Song, G.~Zhang, Q.~Zhang, S.~Liu, J.~Hou, Y.~Xin, M.~Yuan, G.~Zhang \emph{et~al.}, ``A survey of the development of quadruped robots: Joint configuration, dynamic locomotion control method and mobile manipulation approach,'' \emph{Biomimetic Intelligence and Robotics}, vol.~2, no.~1, p. 100029, 2022.

\bibitem{bertoncelli2020linear}
F.~Bertoncelli, F.~Ruggiero, and L.~Sabattini, ``Linear time-varying mpc for nonprehensile object manipulation with a nonholonomic mobile robot,'' in \emph{IEEE international conference on robotics and automation (ICRA)}, 2020, pp. 11\,032--11\,038.

\bibitem{polverini2020multi}
M.~P. Polverini, A.~Laurenzi, E.~M. Hoffman, F.~Ruscelli, and N.~G. Tsagarakis, ``Multi-contact heavy object pushing with a centaur-type humanoid robot: Planning and control for a real demonstrator,'' \emph{IEEE Robotics and Automation Letters}, vol.~5, no.~2, pp. 859--866, 2020.

\bibitem{boston_dynamics_spot_arm}
B.~Dynamics, ``Spot arm,'' \url{https://bostondynamics.com/products/spot/arm/}, accessed: 2024-09-07.

\bibitem{zhou2024design}
J.~Zhou, X.~Song, and K.~Karydis, ``Design of an end-effector with application to avocado harvesting,'' in \emph{IEEE International Conference on Advanced Intelligent Mechatronics (AIM)}, 2024, pp. 1241--1246.

\bibitem{bircher2016receding}
A.~Bircher, M.~Kamel, K.~Alexis, H.~Oleynikova, and R.~Siegwart, ``Receding horizon" next-best-view" planner for 3d exploration,'' in \emph{IEEE international conference on robotics and automation (ICRA)}, 2016, pp. 1462--1468.

\bibitem{dunn2009developing}
E.~Dunn, J.~Van Den~Berg, and J.-M. Frahm, ``Developing visual sensing strategies through next best view planning,'' in \emph{IEEE/RSJ International Conference on Intelligent Robots and Systems}, 2009, pp. 4001--4008.

\bibitem{breyer2022closed}
M.~Breyer, L.~Ott, R.~Siegwart, and J.~J. Chung, ``Closed-loop next-best-view planning for target-driven grasping,'' in \emph{IEEE/RSJ International Conference on Intelligent Robots and Systems (IROS)}, 2022, pp. 1411--1416.

\bibitem{manipulation}
\BIBentryALTinterwordspacing
R.~Tedrake, \emph{Robotic Manipulation}, 2024. [Online]. Available: \url{http://manipulation.mit.edu}
\BIBentrySTDinterwordspacing

\bibitem{bdaiinstitute_spot_ros2}
B.~Lab, ``spot\_ros2: A collection of ros 2 nodes for the boston dynamics spot robot,'' \url{https://github.com/bdaiinstitute/spot_ros2}, accessed: 2024-09-07.

\bibitem{liu2018self}
Y.~Liu, C.~Yu, J.~Sheng, and T.~Zhang, ``Self-collision avoidance trajectory planning and robust control of a dual-arm space robot,'' \emph{International Journal of Control, Automation and Systems}, vol.~16, no.~6, pp. 2896--2905, 2018.

\bibitem{lei2020real}
M.~Lei, T.~Wang, C.~Yao, H.~Liu, Z.~Wang, and Y.~Deng, ``Real-time kinematics-based self-collision avoidance algorithm for dual-arm robots,'' \emph{Applied Sciences}, vol.~10, no.~17, p. 5893, 2020.

\bibitem{moveit_ai}
\BIBentryALTinterwordspacing
{MoveIt AI}, ``Moveit ai,'' accessed: 2024-09-15. [Online]. Available: \url{https://moveit.ai/}
\BIBentrySTDinterwordspacing

\bibitem{ultralytics2023yolov8}
Ultralytics, ``Ultralytics yolov8,'' \url{https://github.com/ultralytics/ultralytics}, 2023, accessed: 2024-09-07.

\bibitem{open3d}
\BIBentryALTinterwordspacing
{Open3D}, ``Open3d: A modern library for 3d data processing,'' accessed: 2024-09-15. [Online]. Available: \url{https://www.open3d.org}
\BIBentrySTDinterwordspacing

\bibitem{xgrids_lixell2}
\BIBentryALTinterwordspacing
{Xgrids}, ``Lixell2,'' accessed: 2024-09-15. [Online]. Available: \url{https://xgrids.com/lixell2}
\BIBentrySTDinterwordspacing

\bibitem{teng2023multimodal}
H.~Teng, Y.~Wang, X.~Song, and K.~Karydis, ``Multimodal dataset for localization, mapping and crop monitoring in citrus tree farms,'' in \emph{International Symposium on Visual Computing}, 2023, pp. 571--582.

\bibitem{liu2022safely}
Z.~Liu, C.~Mucchiani, K.~Ye, and K.~Karydis, ``Safely catching aerial micro-robots in mid-air using an open-source aerial robot with soft gripper,'' \emph{Frontiers in Robotics and AI}, vol.~9, p. 1030515, 2022.

\bibitem{liu2020sorx}
Z.~Liu, Z.~Lu, and K.~Karydis, ``Sorx: A soft pneumatic hexapedal robot to traverse rough, steep, and unstable terrain,'' in \emph{IEEE International Conference on Robotics and Automation (ICRA)}, 2020, pp. 420--426.

\bibitem{liu2021position}
Z.~Liu and K.~Karydis, ``Position control and variable-height trajectory tracking of a soft pneumatic legged robot,'' in \emph{IEEE/RSJ International Conference on Intelligent Robots and Systems (IROS)}, 2021, pp. 1708--1709.

\end{thebibliography}

\end{document}